\documentclass{article}

\usepackage[nonatbib, preprint]{neurips_2025}

\usepackage[utf8]{inputenc} 
\usepackage[T1]{fontenc}    
\usepackage[numbers]{natbib} 
\usepackage{booktabs}       
\usepackage{amsfonts}       
\usepackage{nicefrac}       
\usepackage{microtype}      
\usepackage{xcolor}         
\usepackage{graphicx}       
\usepackage{amsmath}
\usepackage{soul}
\usepackage{xcolor}
\usepackage{tikz}
\usepackage{subcaption}
\usetikzlibrary{calc, positioning}

\newcommand{\target}[1]{\sethlcolor{green!40}\hl{#1}}  
\newcommand{\premise}[1]{\sethlcolor{yellow!40}\hl{#1}}
\newcommand{\contrad}[1]{\sethlcolor{red!40}\hl{#1}}  
\newcommand{\fig}[1]{Figure~#1}
\newcommand{\savespace}{\vspace{-1em}}

\title{A Straightforward Pipeline for Targeted Entailment and Contradiction Detection}

\author{%
  Antonin Sulc \\
  LBNL \\
  \texttt{asulc@lbl.gov}
}

\begin{document}

\maketitle

\begin{abstract}
Finding the relationships between sentences in a document is crucial for tasks like fact-checking, argument mining, and text summarization. A key challenge is to identify which sentences act as premises or contradictions for a specific claim. Existing methods often face a trade-off: transformer attention mechanisms can identify salient textual connections but lack explicit semantic labels, while Natural Language Inference (NLI) models can classify relationships between sentence pairs but operate independently of contextual saliency. 

In this work, we introduce a method that combines the strengths of both approaches for a targeted analysis. Our pipeline first identifies candidate sentences that are contextually relevant to a user-selected target sentence by aggregating token-level attention scores. 
It then uses a pretrained NLI model to classify each candidate as a premise (entailment) or contradiction. By filtering NLI-identified relationships with attention-based saliency scores, our method efficiently isolates the most significant semantic relationships for any given claim in a text.
\end{abstract}

\section{Introduction}
\savespace
Large-scale text data has increased the need for tools that can help humans understand complex documents efficiently. 
The challenge is not in one of information retrieval, but of sens-making: in an era of sophisticated misinformation and information overload, the ability to critically evaluate a claim is essential. 
Whether analyzing a news article, a scientific paper, a legal contract, or a corporate report, a user's primary goal is often to verify a specific statement. 
This requires untangling the document's argumentative structure to answer fundamental questions: 
(1) What evidence is presented to support this claim? (2) Are there any statements within the text that contradict it? Manually performing this analysis is time-consuming and prone to error, creating a pressing need for automated tools that facilitate targeted critical analysis.

Large Language Models (LLMs) have demonstrated capabilities in text understanding. The self-attention mechanism, a core component of their architecture \cite{vaswani2017attention}. It is particularly tailored to capture dependencies between tokens. The resulting attention maps can be interpreted as a measure of token-level saliency, indicating which parts of a text are important for understanding other parts. 
However, these attention scores lack explicit semantic meaning: high attention between two phrases does not, on its own, tell us if one supports or contradicts the other.

\begin{figure*}[t]
\centering 
\resizebox{1.0\linewidth}{!}{
\begin{tikzpicture}[
    text_box/.style={
        text width=6.5cm, 
        align=left,
        font=\linespread{0.9}\selectfont
    },
    arrow_style/.style={
        -latex,
        thick,
        rounded corners,
    }
]

\node[text_box] (input_text) {
    \textbf{Input Document} \\
    \protect\target{The ocean is a crucial carbon sink.} 
    Marine ecosystems absorb vast amounts of CO2. 
    Conversely, the ocean does not play any role in climate regulation. Protecting these habitats is vital for mitigating climate change. This is because healthy oceans sequester carbon.
};

\node[text_box, right=2.5cm of input_text] (output_text) {
    \textbf{Annotated Output} \\
    \protect\target{The ocean is a crucial carbon sink.} 
    \protect\premise{Marine ecosystems absorb vast amounts of CO2.} 
    \protect\contrad{Conversely, the ocean does not play any role in climate regulation.} Protecting these habitats is vital for mitigating climate change. This is because healthy oceans sequester carbon.
};

\draw[arrow_style] (input_text.east) -- (output_text.west)
    node[midway, above, font=\scriptsize\bfseries, text=blue!60!black] {LLM+NLI}
    node[midway, below, font=\tiny] {};
\end{tikzpicture}}
\caption{Given an input with a user-selected \textbf{target} sentence (green), our method uses attention saliency and NLI to automatically identify and highlight the corresponding \textbf{premise} (yellow) and \textbf{contradiction} (red).}
\label{fig:teaser}
\vspace{-1.5em}
\end{figure*}
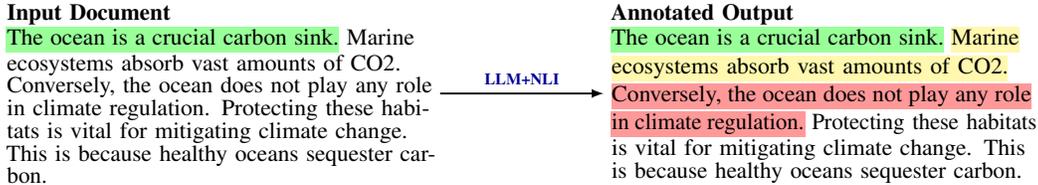

While no existing work offers a direct, integrated solution for our targeted analysis task, related research falls into three main categories, each with significant limitations. The first is the field of Natural Language Inference (NLI), which has produced models specifically trained to recognize semantic relationships between a sentence pair (a \textit{premise} and a \textit{hypothesis}) \citep{williams2017broad}. Models like \texttt{facebook/bart-large-mnli}~\cite{lewis2019bart} can classify these pairs with high accuracy. The primary drawback of pure NLI is its context-agnostic nature and computational inefficiency. To find relationships for a target sentence, one would need to compare it against every other sentence in the document, a process with quadratic complexity ($O(n^2)$) that is impractical for long texts. 
More importantly, this approach treats all sentence pairs equally, failing to distinguish between relationships that are central to the document's discourse and those that are trivial or incidental.

A second related area is Argument Mining, which aims to automatically extract argumentative components (e.g., claims, premises) and their relations (e.g., support, contradiction) from text \cite{stab2017parsing}. While conceptually similar, argument mining systems are often complex, heavyweight solutions that require extensive, domain-specific annotated data for training. They are typically designed for structured, formal argumentation and may not be easily adaptable as a lightweight, general-purpose tool for interactive text exploration. Their complexity stands in contrast to our goal of a straightforward, model-agnostic pipeline.

Finally, a third line of work involves the interpretation and visualization of attention mechanisms themselves \cite{vig2019multiscale}. These studies provide valuable insights into how LLMs process information but are primarily diagnostic in nature. They reveal the existence of connections but do not assign explicit semantic labels to them. A high attention score signals relevance but fails to specify the nature of that relevance, be it support, contradiction, elaboration, or simple co-reference.

This work addresses the gap left by these approaches. We propose a two-stage method that integrates causal LLM attention with NLI to create a system that is both context-aware and semantically explicit. Our core contribution is the idea that a meaningful semantic relationship between a target sentence $S_{target}$ and another sentence $S_j$ should be characterized by both: (1) a high degree of mutual attention, indicating contextual saliency, and (2) a clear "entailment" or "contradiction" label from an NLI model. By combining these signals, our method efficiently filters for relevant premises and contradictions, enabling a practical and targeted analysis of any claim within a document.

\savespace
\section{Method}
\savespace
To map the logical structure surrounding a specific claim in a document, our method operates on a simple yet powerful principle: a significant relationship between a target sentence and another sentence must be both \textbf{contextually salient} and \textbf{semantically explicit}. Relying on one signal alone is insufficient. Raw attention scores from a language model, while indicative of textual connections, are often noisy and only show that two passages are related, but not \textit{how}. Conversely, applying an NLI model to compare a target sentence against all others is computationally expensive and context-agnostic.

Our pipeline integrates these two signals to overcome their individual limitations. It is a three-step process for analyzing a chosen target sentence, $S_{target}$: (1) we use attention as a wide-angle spotlight to efficiently find potentially important connections to $S_{target}$, (2) we use NLI as a precision lens to classify their meaning relative to $S_{target}$, and (3) we fuse these signals to render a final verdict on the core premises and contradictions.

\savespace
\subsection{Identifying Contextual Saliency with Attention}
\savespace
The first step is to efficiently identify which sentences are worth comparing to our target sentence, $S_{target}$. The attention mechanism of a causal language model provides a robust signal for this task. We extract attention weights from the \textbf{final layer} of the transformer, as these layers are known to encode more abstract and semantic dependencies crucial for understanding discourse structure \cite{Jawahar2019BERTLayers}.

The process begins by feeding the entire document into a pre-trained language model and extracting the token-to-token attention matrix, $A_{tok}$ of the entire document. To create a sentence-level view, we aggregate these token-level scores with sentence tokenizer~\cite{bird2006nltk}. The saliency of sentence $S_j$ relative to sentence $S_i$ is defined as the mean attention score from all tokens in $S_i$ to all tokens in $S_j$:
\begin{equation}
    A_{sent}[i, j] = \frac{1}{|S_i| |S_j|} \sum_{k \in S_i} \sum_{l \in S_j} A_{tok}[k, l]
\end{equation}
This yields a saliency matrix, $A_{sent}$, which serves as a weighted, directed graph. To find candidate sentences for our target, $S_{target}$, we select all other sentences $S_j$ where $A_{sent}[\text{target}, j]$ exceeds a certain threshold, indicating a strong, contextually-aware connection that warrants further semantic analysis.

\savespace
\subsection{Classifying Semantic Relationships with NLI}
\savespace

While the saliency matrix tells us \textbf{where} to look, it does not tell us \textbf{what} we are seeing. The second stage addresses this by assigning precise semantic labels to the filtered candidate sentences. For this, we employ a pre-trained NLI model.

For each candidate sentence $S_c$ identified in the previous step, we form two ordered pairs with our target sentence, $S_{target}$, to check for both premises and contradictions:

\begin{enumerate}
    \item \textbf{To find premises}: We treat the candidate as the premise and the target as the hypothesis, $(S_c, S_{target})$. If the NLI model classifies this pair as `entailment`, we label $S_c$ as a \textbf{premise} for our target.
    \item \textbf{To find contradictions}: We treat the target as the premise and the candidate as the hypothesis, $(S_{target}, S_c)$. If the model returns `contradiction`, we label $S_c$ as a \textbf{contradiction}.
\end{enumerate}
Pairs classified as `neutral` are discarded. This stage produces a relationship matrix, $R$, which exhaustively labels every potential semantic link to the target sentence.

\savespace
\subsection{Combining Saliency and Semantics}
\savespace

The final step confirms the significance of the identified relationships. The NLI process, even on filtered candidates, can sometimes flag logically valid but contextually weak connections. The attention-based saliency scores provide the necessary final filter.

We formalize our rule: a sentence $S_c$ is confirmed as a premise or contradiction of $S_{target}$ only if it receives the appropriate NLI label from stage two \textbf{and} its saliency score $A_{sent}[\text{target}, c]$ surpasses a predetermined threshold. This fusion ensures that our final output only contains relationships to the target sentence that are both semantically unambiguous and validated as important by the language model's contextual understanding of the entire document.
\savespace
\section{Experiments and Results}
\savespace

To show the plausibility and effectiveness of our targeted analysis, we designed three test cases with increasing complexity. For each case, we selected a target sentence and applied our pipeline to identify its corresponding premises and contradictions within the text.

To validate our pipeline and underscore its accessibility, all experiments were performed on a conventional laptop (Dell XPS with an Intel i7 processor and 16 GB of RAM), deliberately avoiding the need for specialized GPU hardware. For the attention-based saliency analysis described in we utilized the \texttt{Qwen/Qwen3-1.7B}~\cite{yang2025qwen3}. Following this, for the semantic classification stage, we evaluated several publicly available NLI models. We empirically determined that the \texttt{MoritzLaurer/DeBERTa-v3-large-mnli-fever-anli-ling-wanli}~\cite{laurer2024less} model delivered the most robust and accurate performance across our diverse test cases.

\savespace
\paragraph{Case 1: Direct Factual Relationships.} We first analyzed a short text containing a simple factual claim to test the system's ability to identify direct support and contradiction. The target sentence selected was "\textit{The sun is a star.}"

\begin{figure}[h]
  \centering
  \target{The sun is a star.} \premise{It is the center of our solar system.} \contrad{The sun is a planet.} All planets revolve around it.
  \caption{Analysis of the target sentence (green). The system correctly identifies one premise (yellow) and one contradiction (red) that are both semantically valid and contextually salient.}
  \label{fig:example1}
\end{figure}

\savespace
As shown in \fig{fig:example1}, our system correctly identified two key relationships. The sentence "\textit{It is the center of our solar system}" was flagged as a premise (entailment), while "\textit{The sun is a planet}" was flagged as a contradiction. Both sentences exhibited high attention scores relative to the target, allowing them to pass the saliency filter and be confirmed by the NLI model. Specifically, the attention between the target and the contradiction was 0.1288, significantly above the text's average inter-sentence attention of 0.0349.

\savespace
\paragraph{Case 2: Implied Contradiction.} Next, we crafted a narrative text where a contradiction arises from a logical inconsistency rather than a direct factual opposition. The target sentence was the initial statement of intent: "\textit{Mark decided to build a bookshelf from scratch.}"

\begin{figure}[h]
\centering
\parbox{0.9\linewidth}{
\target{Mark decided to build a bookshelf from scratch.} He started by carefully measuring the space in his living room. Next, he bought high-quality oak wood and cut each piece to the exact size. He spent a full weekend sanding, assembling, and staining the bookshelf. \contrad{He found that IKEA fits perfectly to his requirements.} In the end, the bookshelf was sturdy, fit perfectly in the space, and looked professionally made.
}
\caption{Detection of an implied contradiction. The outcome (finding an IKEA solution) logically contradicts the initial intention to build from scratch (target, green).}
\label{fig:example2}
\end{figure}

The system successfully identified the subtle contradiction shown in \fig{fig:example2}. The statement "\textit{He found that IKEA fits perfectly to his requirements}" logically undermines the initial goal. This relationship was captured because the attention score between the two sentences (0.0768) was a standout, more than double the standard deviation (0.0284) above the mean attention (0.0135) for this text. This demonstrates the model's ability to connect distant but semantically opposed concepts.

\savespace
\paragraph{Case 3: Complex Argument Analysis.} Finally, we analyzed a more nuanced text involving a central claim supported by evidence and challenged by a counter-argument. We targeted the sentence presenting specific evidence: "\textit{In fact, several pilot programs have reported savings of up to 60\% on lighting expenses after switching to LEDs.}"

\begin{figure}[h]
  \centering
  \parbox{0.9\linewidth}{
     Many cities are exploring the idea of replacing traditional streetlights with smart LED systems. \premise{These smart lights are far more energy-efficient than conventional bulbs, helping municipalities cut electricity costs.} \target{In fact, several pilot programs have reported savings of up to 60\% on lighting expenses after switching to LEDs.} However, the data systems that control these lights require regular software updates and cybersecurity measures, which have added unexpected ongoing costs for some cities. \contrad{In some cases, these challenges have led municipalities to abandon LED upgrades altogether and return to conventional lighting.}}
  \caption{A unified analysis of a complex argument. For the target evidence (green), the system identifies the broader claim it supports as a premise (yellow) and a counter-argument that challenges its implications as a contradiction (red).}
  \label{fig:example3}
\end{figure}

\savespace
The results in \fig{fig:example3} showcase the system's ability to parse a multi-faceted argument. When analyzing the target evidence, it correctly identified the preceding general claim ("\textit{These smart lights are far more energy-efficient...}") as its premise. The attention score between this pair (0.0024) was sufficient to proceed to NLI, which confirmed the entailment. Furthermore, it identified the sentence about cities abandoning the upgrades as a contradiction. Although its attention score was lower (0.0008), it was still deemed salient enough in the local context to be evaluated, and the NLI model confirmed its contradictory nature relative to the successful pilot programs.

\savespace
\section{Discussion and Future Work}
\savespace

The results from our test cases suggest that this two-staged approach is a promising direction for interactive text analysis tools. By allowing users to select a target sentence, our method enables focused and meaningful exploration of a document's argumentative structure.

By combining contextual saliency from attention with explicit classification from NLI models, our method offers a more robust analysis than either technique alone. The attention filter narrows the search space to sentence pairs the model deems interconnected, while NLI labels provide the semantic grounding that raw attention scores lack. An additional advantage is that our approach is not tied to any specific high-quality LLM; since it relies on the attention mechanism common to transformer models, it can be applied broadly and will likely improve automatically as LLM capabilities advance.

While experimenting, we identified two main limitations. (1) Its performance depends on the quality of the underlying models: attention patterns may vary, and models not tuned for structured reasoning can produce noisy saliency maps. (2) NLI models, while accurate, still struggle with nuance, sarcasm, or complex syntax. Our aggregation of token-level attention to the sentence level via arithmetic mean is also a simplification; more sophisticated methods could yield more precise saliency scores. Future work should explore user-centric design, enabling users to adjust the saliency threshold to control analysis granularity.

\clearpage
\bibliographystyle{plainnat}
\bibliography{references}

\begin{thebibliography}{9}
\providecommand{\natexlab}[1]{#1}
\providecommand{\url}[1]{\texttt{#1}}
\expandafter\ifx\csname urlstyle\endcsname\relax
  \providecommand{\doi}[1]{doi: #1}\else
  \providecommand{\doi}{doi: \begingroup \urlstyle{rm}\Url}\fi

\bibitem[Bird(2006)]{bird2006nltk}
Steven Bird.
\newblock Nltk: the natural language toolkit.
\newblock In \emph{Proceedings of the COLING/ACL 2006 interactive presentation
  sessions}, pages 69--72, 2006.

\bibitem[Jawahar et~al.(2019)Jawahar, Sagot, and Seddah]{Jawahar2019BERTLayers}
Ganesh Jawahar, Beno{\^i}t Sagot, and Djam{\'e} Seddah.
\newblock What does bert learn about the structure of language?
\newblock In \emph{ACL}, 2019.
\newblock URL \url{https://aclanthology.org/P19-1356.pdf}.

\bibitem[Laurer et~al.(2024)Laurer, Van~Atteveldt, Casas, and
  Welbers]{laurer2024less}
Moritz Laurer, Wouter Van~Atteveldt, Andreu Casas, and Kasper Welbers.
\newblock Less annotating, more classifying: Addressing the data scarcity issue
  of supervised machine learning with deep transfer learning and bert-nli.
\newblock \emph{Political Analysis}, 32\penalty0 (1):\penalty0 84--100, 2024.

\bibitem[Lewis et~al.(2019)Lewis, Liu, Goyal, Ghazvininejad, Mohamed, Levy,
  Stoyanov, and Zettlemoyer]{lewis2019bart}
Mike Lewis, Yinhan Liu, Naman Goyal, Marjan Ghazvininejad, Abdelrahman Mohamed,
  Omer Levy, Ves Stoyanov, and Luke Zettlemoyer.
\newblock Bart: Denoising sequence-to-sequence pre-training for natural
  language generation, translation, and comprehension.
\newblock \emph{arXiv preprint arXiv:1910.13461}, 2019.

\bibitem[Stab and Gurevych(2017)]{stab2017parsing}
Christian Stab and Iryna Gurevych.
\newblock Parsing argumentation structures in persuasive essays.
\newblock \emph{Computational Linguistics}, 43\penalty0 (3):\penalty0 619--659,
  2017.

\bibitem[Vaswani et~al.(2017)Vaswani, Shazeer, Parmar, Uszkoreit, Jones, Gomez,
  Kaiser, and Polosukhin]{vaswani2017attention}
Ashish Vaswani, Noam Shazeer, Niki Parmar, Jakob Uszkoreit, Llion Jones,
  Aidan~N Gomez, {\L}ukasz Kaiser, and Illia Polosukhin.
\newblock Attention is all you need.
\newblock \emph{Advances in neural information processing systems}, 30, 2017.

\bibitem[Vig(2019)]{vig2019multiscale}
Jesse Vig.
\newblock A multiscale visualization of attention in the transformer model.
\newblock \emph{arXiv preprint arXiv:1906.05714}, 2019.

\bibitem[Williams et~al.(2017)Williams, Nangia, and Bowman]{williams2017broad}
Adina Williams, Nikita Nangia, and Samuel~R Bowman.
\newblock A broad-coverage challenge corpus for sentence understanding through
  inference.
\newblock \emph{arXiv preprint arXiv:1704.05426}, 2017.

\bibitem[Yang et~al.(2025)Yang, Li, Yang, Zhang, Hui, Zheng, Yu, Gao, Huang,
  Lv, et~al.]{yang2025qwen3}
An~Yang, Anfeng Li, Baosong Yang, Beichen Zhang, Binyuan Hui, Bo~Zheng, Bowen
  Yu, Chang Gao, Chengen Huang, Chenxu Lv, et~al.
\newblock Qwen3 technical report.
\newblock \emph{arXiv preprint arXiv:2505.09388}, 2025.

\end{thebibliography}

\end{document}